\documentclass[a4paper, 10pt, conference]{ieeeconf}      
\usepackage{FG2025}
\usepackage{amsmath}
\usepackage{graphicx}
\usepackage{amssymb}

\FGfinalcopy 

\IEEEoverridecommandlockouts                              
\def\FGPaperID{242} 

\title{\LARGE \bf
Wearable-Derived Behavioral and Physiological Biomarkers for Classifying Unipolar and Bipolar Depression Severity}

\author{\parbox{16cm}{\centering
    {\large Yassine Ouzar$^{1,2}$, Clémence Nineuil$^{2}$, 
    Fouad Boutaleb$^{1,2}$, Emery Pierson$^{3}$, Ali Amad$^{2}$, and Mohamed Daoudi$^{1,4}$}\\
    {\normalsize
    $^1$ Univ. Lille, CNRS, Centrale Lille, Institut Mines-Télécom, UMR 9189 CRIStAL, F-59000 Lille, France\\
    $^2$Univ. Lille, Inserm, CHU Lille, U1172 - LilNCog - Lille Neuroscience \& Cognition, F-59000 Lille, France\\
    $^3$ LIX, École Polytechnique, IPP Paris\\
    $^4$ IMT Nord Europe, Institut Mines-Télécom, Univ. Lille, Centre for Digital Systems, F-59000 Lille, France\\
    }}
    \thanks{The French State under the France-2030 programme and the Initiative of Excellence of the University of Lille are acknowledged for the funding and support granted to the R-CDP-24-005-CALYPSO project.}
}

\begin{document}

\maketitle

\begin{abstract}

Depression is a complex mental disorder characterized by a diverse range of observable and measurable indicators that go beyond traditional subjective assessments. Recent research has increasingly focused on objective, passive, and continuous monitoring using wearable devices to gain more precise insights into the physiological and behavioral aspects of depression. However, most existing studies primarily distinguish between healthy and depressed individuals, adopting a binary classification that fails to capture the heterogeneity of depressive disorders.
In this study, we leverage wearable devices to predict depression subtypes—specifically unipolar and bipolar depression—aiming to identify distinctive biomarkers that could enhance diagnostic precision and support personalized treatment strategies. To this end, we introduce the CALYPSO dataset, designed for non-invasive detection of depression subtypes and symptomatology through physiological and behavioral signals, including blood volume pulse, electrodermal activity, body temperature, and three-axis acceleration. Additionally, we establish a benchmark on the dataset using well-known features and standard machine learning methods.
Preliminary results indicate that features related to physical activity, extracted from accelerometer data, are the most effective in distinguishing between unipolar and bipolar depression, achieving an accuracy of 96.77\%. Temperature-based features also showed high discriminative power, reaching an accuracy of 93.55\%. These findings highlight the potential of physiological and behavioral monitoring for improving the classification of depressive subtypes, paving the way for more tailored clinical interventions.

\end{abstract}

\begin{figure*}[h]
\begin{center}
\includegraphics[scale=0.36]{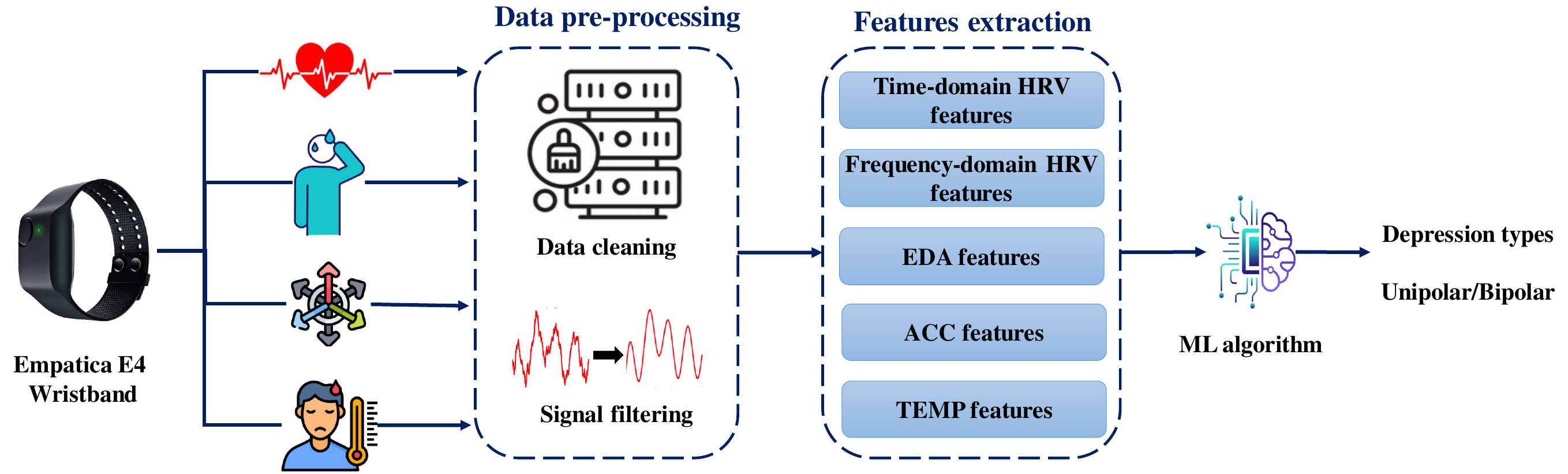}

\end{center}
   \caption{Overview of the proposed framework for classifying depression types based on physiological signals collected from the Empatica E4 wristband. The initial step involves pre-processing the raw sensor data to remove noise and artifacts, thereby enhancing the signal-to-noise ratio. Subsequently, time-domain, frequency-domain, and statistical features are extracted from the filtered signals. Finally, these extracted features are fed individually and combined to ML algorithms for the classification of unipolar and bipolar depression.
}
\label{framework}
\end{figure*}

\section{INTRODUCTION}

Depression is a common psychiatric disorders affecting approximately 280 million people worldwide, accounting for 3.8\% of the global population. Its prevalence is higher among women and individuals over the age of 60 \cite{world2023depressive}. The main symptoms of depression include persistent feelings of sadness, a loss of interest or pleasure in daily activities, sleep disturbances such as insomnia, an inability to experience joy, and, in severe cases, suicidal thoughts \cite{liu2020changes}.
Although cultural, psychological, and biological factors are recognized as contributing to depression, and effective treatments such as psychological therapies and medications exist, the complex underlying pathogenesis remains poorly understood \cite{liu2020changes}. A significant barrier to progress lies in the reliance on subjective assessment tools, such as self-reported questionnaires and clinician-administered scales. These methods are prone to biases and the retrospective nature of the information collected \cite{walschots2024using}.

To address these limitations, there is growing interest in objective assessment methods, particularly those utilizing wearable devices \cite{rykov2021digital, pedrelli2020monitoring}. Wearable technology offers several advantages for depression assessment and monitoring, including continuous real-time data collection, improved ecological validity, and the ability to measure objective biomarkers. 
On the other hand, prior research have focused mainly on discriminate between healthy and depressed individuals. However, depression is a heterogeneous condition characterized by diverse clinical subtypes and categories such as unipolar and bipolar disorder, each with distinct behavioral, physiological, and neurobiological signatures \cite{drysdale2017resting}. 
Unipolar depression, often referred to as major depressive disorder, is characterized by persistent depressed mood, loss of interest, recurrent thoughts of death, and a range of physical and cognitive symptoms, typically occurring without episodes of mania \cite{marx2023major}. In contrast, bipolar depression shares many overlapping symptoms with unipolar depression but is distinguished by the presence of alternating episodes of depression and mania or hypomania \cite{mcintyre2020bipolar}. Bipolar patients may also exhibit atypical features, such as hypersomnia, hyperphagia, and psychotic features, which are less common in unipolar depression \cite{mitchell2008diagnostic}. This distinction highlights the complexity and heterogeneity of depressive disorders. Therefore, simply distinguishing between healthy and depressed overlooks the important variations within depression that can significantly impact clinical outcomes and treatment approaches.

In light of the limitations of previous studies, particularly their reliance on subjective assessments and limited exploration of depression subtypes, we aimed to investigate the potential of objective markers derived from wearable device in discriminating between unipolar and bipolar depression. This study is, to our knowledge, the first to specifically focus on detecting depression subtypes, moving beyond the typical binary distinction of healthy vs. depressed individuals.

The main contributions of this study are summarized as follows:

\begin{enumerate}

\item {We introduce a novel dataset named CALYPSO depression, specifically designed to detect various types of depression and their characteristics. Unlike most existing datasets, CALYPSO focuses on objective markers extracted from wearable devices, including HRV, EDA, temperature, and accelerometer data, alongside behavioral signals such as vocal tone, facial expressiveness, and head movements.}

\item {We establish a benchmark for differentiating between unipolar and bipolar depression using a comprehensive set of time-domain, frequency-domain, and statistical features derived from Empatica wearable sensor data.}

\item {We evaluated the extracted features, both individually and in combination, using several commonly used machine learning algorithms.}

\end{enumerate}

\section{RELATED WORK}

By surveying existing research on depression, two primary diagnostic approaches can be identified, each differing in methodology and focus. Traditional approaches heavily rely on subjective assessments, such as patient self-reports, structured interviews, and clinician observations guided by DSM-5 criteria \cite{american2000diagnostic}. Although commonly used in clinical practice, this method has inherent limitations, including susceptibility to biases, inconsistencies in patient reporting, and reliance on retrospective information \cite{walschots2024using}. Moreover, the heterogeneous nature of depression poses significant challenges for accurate diagnosis \cite{kim2018automatic}. In contrast, the second approach highlights the need for more objective and comprehensive diagnostic approaches that go beyond subjective assessments, integrating quantifiable biomarkers spanning behavioral, physiological, and neurobiological domains \cite{pedrelli2020monitoring}. These include alterations in physiological responses, reduced physical activity, changes in vocal characteristics such as tone and prosody, and decreased facial expressiveness and emotional reactivity, among other indicators. 

Behavioral markers such as facial expressions, head movements, and vocal prosody are valuable indicators of affective states, including depression \cite{islam2024facepsy}. However, their reliability can be influenced by various factors, such as social and cultural differences, situational context, and environmental conditions (e.g., camera angles, lighting, ambiguity, and noise). Additionally, individuals can consciously control or mask their expressions, further complicating the interpretation of these markers \cite{ouzar2023reconnaissance}.
On the other hand, physiological signals stand out as particularly important and promising. Signals such as heart rate variability (HRV), electrodermal activity (EDA), and respiratory patterns offer several key advantages over other markers. They directly reflect autonomic nervous system (ANS) activity, which is often dysregulated in individuals experiencing stress or emotional disturbances linked to depression \cite{vinkers2021integrated}. Unlike subjective assessments or behavioral markers, physiological signals provide objective, quantifiable measurements that are less susceptible to subjective interpretation or patient bias \cite{ouzar2023reconnaissance}. Additionally, they enable passive, continuous, and real-time monitoring via wearable devices, enhancing ecological validity and reducing reliance on retrospective self-reports \cite{de2022digital}.

Wearable devices like wristbands and smartwatches are increasingly used to monitor physiological signals such as heart rate and skin conductance for depression detection \cite{lee2021current, rykov2021digital, walschots2024using}. Most existing studies have examined the use of these physiological cues separately. Hartmann et al. found significant correlations between depression and various heart rate variability features. Specifically, individuals with major depression tended to exhibit reduced high-frequency HRV (HRV-HF), low-frequency HRV (HRV-LF), very low-frequency HRV (HRV-VLF), standard deviation of normal-to-normal intervals (SDNN), and root mean square of successive differences (RMSSD), along with an increased LF/HF ratio when compared to healthy individuals \cite{hartmann2019heart}. Regarding electrodermal activity, studies have consistently shown that patients with depression exhibit lower skin conductance levels during rest compared to healthy control subjects \cite{pedrelli2020monitoring, kim2018automatic}. Physical activity has also been linked to depression. A meta-analysis by Schuch et al. reported that greater physical activity, as measured by accelerometers, was associated with lower rates of depression \cite{schuch2017physical}. Additionally, research on body temperature in depression has yielded interesting results. Tazawa et al. demonstrated that individuals with depression may exhibit elevated skin temperatures compared to healthy individuals \cite{tazawa2020evaluating}.

While these studies have significantly advanced the understanding of physiological distinctions between healthy individuals and those with depression, they predominantly focus on binary classification (i.e., depressed vs. healthy) \cite{kim2018automatic}. To the best of our knowledge, no existing research has explored the potential of wearable-derived physiological and behavioral data to discriminate between categories and subtypes of depression. Identifying physiological markers that distinguish between depression subtypes, such as unipolar, bipolar, melancholic, atypical, or psychotic depression, could provide deeper insights into the underlying mechanisms and support personalized treatment approaches.

\section{MATERIALS AND METHODS}

\subsection{Dataset} \label{dataset}

The data presented herein are derived from the CALYPSO clinical trial. This monocentric pilot study is designed to identify and describe objective markers associated with the severity of depression and its clinical characteristics. The CALYPSO clinical trial has been reviewed and approved by the regional ethics Committee, ensuring adherence to ethical standards and guidelines. All patients provided written informed consent before participation in the study.

In routine psychiatric practice, patients admitted to the CHU of Lille undergo routine diagnostic evaluations conducted by an attending psychiatrist. A standardized clinical evaluation is performed, including a collection of medical history, a structured interview using the Hamilton Depression Rating Scale to assess the severity of depression, and an investigation of the specific characteristics of severe depression according to the DSM-5 criteria (melancholic, catatonic, psychotic, anxious, or seasonal) \cite{american2000diagnostic}. Patients who meet the DSM-5 inclusion criteria for severe depression are invited to participate in a more comprehensive interview, contributing directly to the CALYPSO depression dataset. 

\begin{figure}[htbp]
\begin{center}
\includegraphics[width=8.5cm, height=7.3cm]{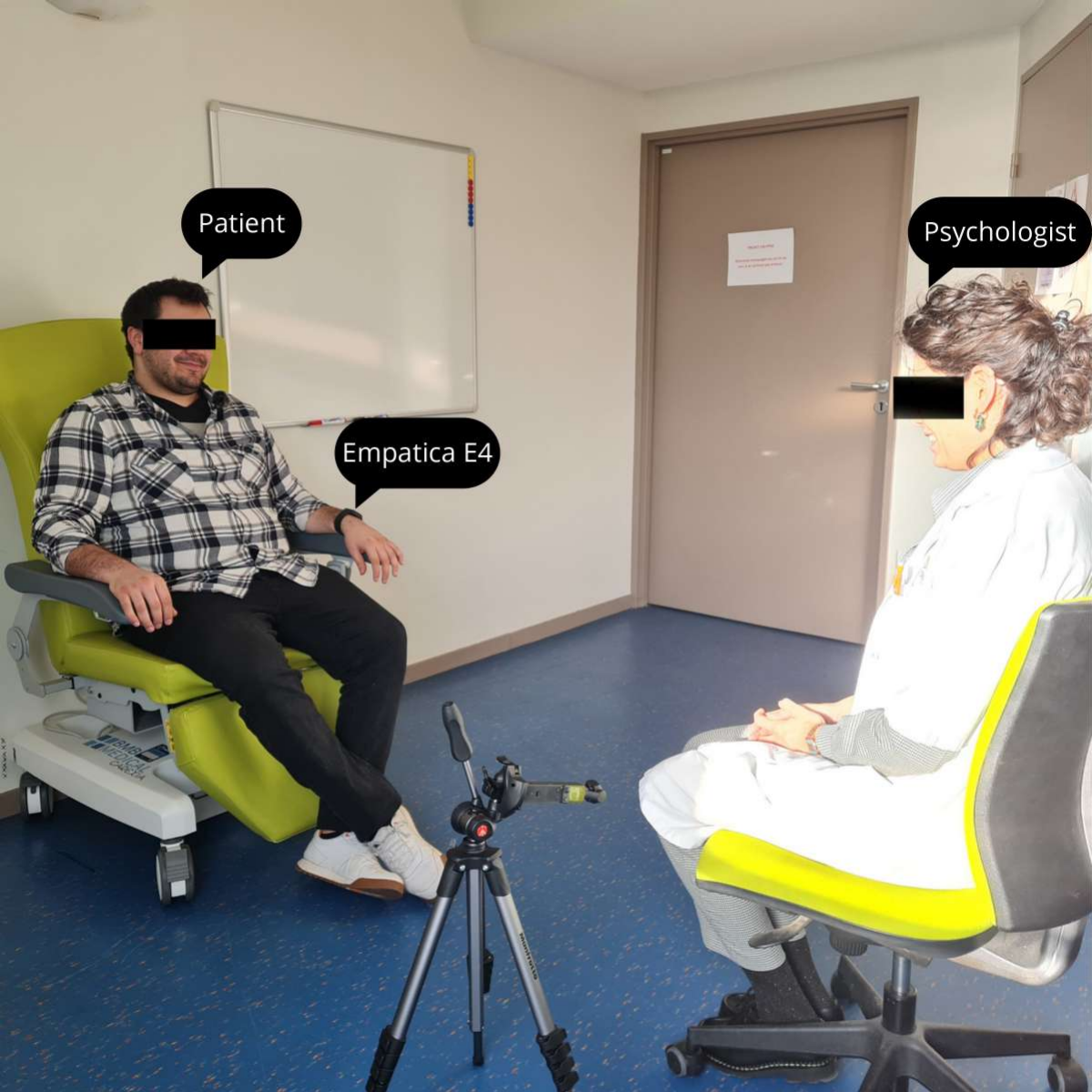}
\end{center}
   \caption{Interview Room Setup for the Calypso Depression Dataset.
}
\label{interview}
\end{figure}

A total of 40 eligible patients diagnosed with severe depression were included in the study. For each participant, an interview was recorded using a camera to capture facial expressions and body movements, and a microphone to analyze vocal features. Additionally, an Empatica E4 wristband was used to monitor physical activity and physiological responses. This study focuses specifically on the signals collected from the Empatica. The participants were predominantly French nationals (ethnicity data were not collected). Each patient took part in a structured clinical interview conducted in a controlled environment, as illustrated in Figure \ref{interview}. The interview consisted of two phases: an initial informal segment involving a brief, non-medical conversation with the clinician, followed by a structured assessment based on the Hamilton Depression Rating Scale (HDRS) to evaluate the severity of depression. On average, the maximum duration of the interviews was 20.53 seconds, with a standard deviation of 8.68 seconds. After the session, the clinician determined the depression subtype and its characteristics based on the patient’s responses and observed behaviors. 

The E4 wristband records six physiological signals in CSV format, including blood volume pulse (BVP), measured using photoplethysmography, average heart rate (HR) computed over 10-second intervals, and interbeat intervals (IBI), which measure the time between heartbeats and provide insights into heart rate variability. It also captures electrodermal activity (EDA), which reflects emotional responses through changes in skin conductance, accelerometer data (ACC), which monitors physical movement and activity, and skin temperature (TEMP), which indicates ANS function. 

The performance of the E4 bracelet has been validated in multiple studies \cite{mccarthy2016validation, chandra2021comparative}, and has been applied in diverse research fields, including sleep monitoring \cite{shi2024sleep}, driving safety \cite{amidei2022driver}, and social stress monitoring \cite{sabour2021ubfc}. Data from 31 participants were available for analysis, as data from 9 participants were excluded due to signal corruption or missing information. Among the six physiological signals recorded by the Empatica E4, only four were used in this study, namely BVP, EDA, ACC, and TEMP. The other two signals were excluded due to the presence of certain corruptions in some recordings and because they are not directly relevant for quantifying depression. It is worth noting that IBI represents the time between successive systolic peaks in the BVP signals, and HR is calculated as 60/IBI to obtain the heart rate in beats per minute \cite{ouzar2022multimodal}.

\subsection{Features extraction} \label{sec::Features extraction}

We describe the set of interpretable features used in this paper. The extracted features include time-domain and frequency-domain heart rate variability (HRV) features derived from BVP signals, alongside features from EDA, ACC, and TEMP data. A detailed description of each feature category is provided below.

\subsubsection{Heart rate variability features}

Heart rate variability is a valuable indicator for analyzing the physiological mechanisms associated with depression. It reflects the balance between the activity of the sympathetic nervous system, linked to stress responses, and the parasympathetic nervous system, associated with relaxation and recovery \cite{pham2021heart}. Research has shown that low HRV is strongly correlated with the presence of depressive disorders \cite{hartmann2019heart, bassett2016literature, gullett2023heart}. 

Two types of HRV features were extracted from BVP signals in time-domain and frequency-domain. Following the pre-processing procedure described in \cite{ouzar2022video}, detrending and a 2nd-order Butterworth band-pass filter were performed to eliminate noise and irrelevant frequency bands while preserving the pulse waveform information. Then, 25 time-domain and 10 frequency- domain HRV features were extracted from the filtered BVP signals using NeuroKit2 toolkit \cite{makowski2021neurokit2}. Details on both are given below.

\paragraph{Time-domain heart rate variability features}

\setlength{\arrayrulewidth}{0.3mm}
\renewcommand{\arraystretch}{1.2}

\begin{table}[htbp]
\centering
\begin{tabular}{|p{2.5cm}|p{5.3cm}|}

\hline
\textbf{HRV Features} & \textbf{Description} \\ \hline
MeanNN & Mean of Normal-to-Normal (NN) Intervals \\ \hline
SDNN & Standard Deviation of NN Intervals \\ \hline
SDANN1, SDANN2, SDANN5 & Standard Deviation of Averages of NN Intervals over 1, 2, \& 5 minutes \\ \hline
SDNNI1, SDNNI2, SDNNI5 & Standard Deviation of NN Intervals within segments of 1, 2, \& 5 minutes \\ \hline
RMSSD & Root Mean Square of Successive Differences between adjacent NN intervals \\ \hline
SDRMSSD & Standard Deviation of RMSSD \\ \hline
SDSD & Standard Deviation of Successive Differences between adjacent NN intervals \\ \hline
CVNN & Coefficient of Variation of NN Intervals \\ \hline
MCVNN & Mean Coefficient of Variation of NN Intervals between adjacent NN intervals \\ \hline
CVSD & Coefficient of Variation of Successive Differences between adjacent NN intervals \\ \hline
IQRNN & Interquartile Range of NN Intervals \\ \hline
MinNN \& MaxNN & Min and Max of NN Intervals \\ \hline
MedianNN & Median of NN Intervals \\ \hline
MADNN & Median Absolute Deviation of NN Intervals \\ \hline

HTI & Heart Rate Turbulence Index \\ \hline
TINN & Triangular Index of NN Intervals \\ \hline
pNN50 \& pNN20 & Percentage of Pairs of Successive NN Intervals \\ \hline
Prc20NN \& Prc80NN & 20th and 80th Percentille of NN Intervals \\ \hline

\end{tabular}
\caption{Time-domain HRV features and their description.}

\label{tab:time_HRV_features}
\end{table}

Time-domain HRV features primarily reflect the beat-to-beat variations in the time intervals between successive heartbeats, known as the IBI or NN (normal-to-normal) intervals. These variations mirror the overall ANS regulation, particularly the balance between the parasympathetic and sympathetic branches. These features provide insights into how the body responds to stress, recovery, and emotional regulation. 25 time-domain HRV features are returned by NeuroKit2 which are described in Table \ref{tab:time_HRV_features}. However, SDANN5 and SDNNI5, calculated from the variability of IBI across 5-minute segments, were excluded due to missing values (NaN) in some recordings.

\paragraph{Frequency-domain heart rate variability features}

\begin{figure}[htbp]
\begin{center}
\includegraphics[width=8.5cm, height=6cm]{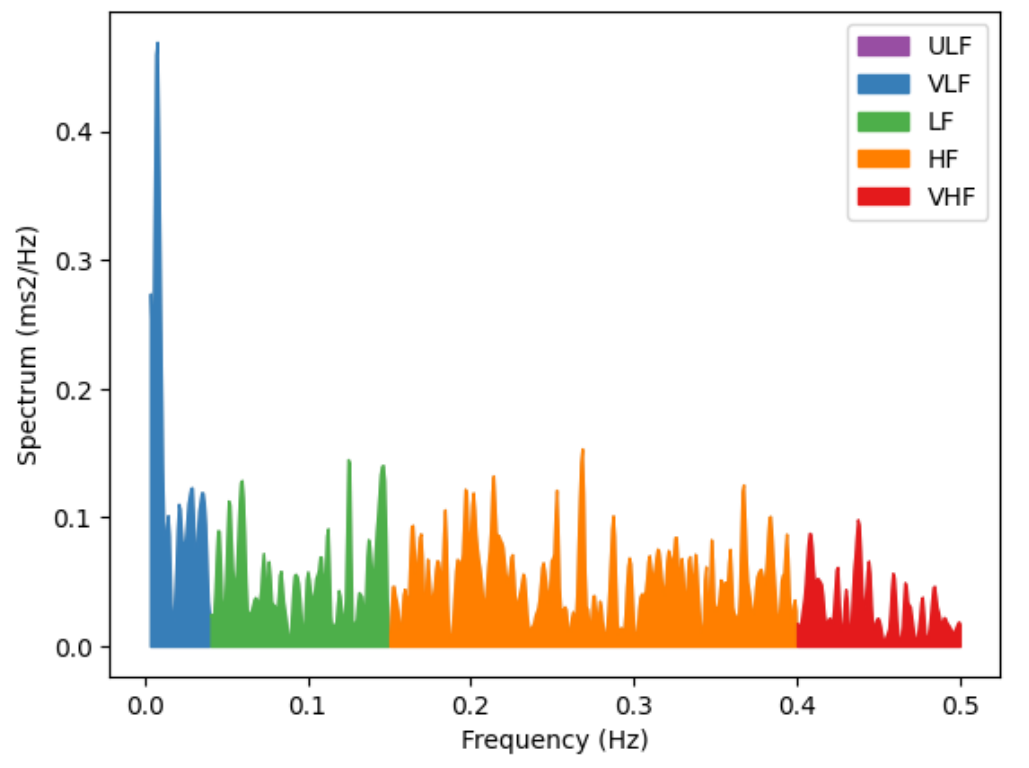}
\end{center}
   \caption{A representative PSD for IBI signal showing The areas of VLF, LF, HF and VHF powers of the HRV.}
\label{fig:HRV}
\end{figure}

Frequency-domain HRV features analyze the periodic oscillations in heart rate at different frequencies, providing insights into the ANS regulation. These features are derived by analyzing the power spectral density of the IBI time series, which decomposes the HRV signal into distinct frequency bands, each associated with different physiological processes. The main frequency bands are the very high frequency (VHF: $\geq 0.4$ Hz), high frequency (HF: $0.15 - 0.4$ Hz), low frequency (LF: $0.04 - 0.15$ Hz), very low frequency (VLF: $0.0033 - 0.04$ Hz), and ultra-low frequency (ULF: $\leq 0.003$ Hz). HRV features in the frequency domain are computed as the area under the PSD curve within the corresponding frequency band. For example, LF and HF powers are computed as the area under the PSD curve corresponding to 0.04-0.15Hz and 0.15-0.4Hz respectively (see Figure \ref{fig:HRV}).

NeuroKit2 returns 10 frequency-domain HRV features, which are detailed in Table \ref{tab:frequency_HRV_features}. Similar to time-domain HRV features, the ULF was excluded due to the presence of missing values. ULF power is typically measured over a long-term segment, often ranging from 24 hours to several hours of continuous IBI data. The time segment should be sufficiently long to capture the very low-frequency oscillations (below 0.0033 Hz), which occur over longer periods, making shorter recordings inadequate for ULF measurement.

\setlength{\arrayrulewidth}{0.3mm}
\renewcommand{\arraystretch}{1.2}

\begin{table}[htbp]
\centering
\begin{tabular}{|p{1.8cm}|p{5.6cm}|}
\hline
\textbf{HRV Features} & \textbf{Description} \\
\hline
TP & Total variance of the HRV signal, reflecting overall autonomic activity \\
\hline
ULF & Power in the ultra low-frequency range (less than 0.0033 Hz), often associated with long-term regulatory mechanisms \\
\hline
VLF & Power in the very low-frequency range (0.0033 to 0.04 Hz), reflecting long-term regulatory mechanisms \\
\hline
LF & Power in the low-frequency range (0.04 to 0.15 Hz), associated with both sympathetic and parasympathetic activity \\
\hline
HF & Power in the high-frequency range (0.15 to 0.4 Hz), primarily reflecting parasympathetic activity (vagal). \\
\hline
VHF & Power in the very high-frequency range (above 0.4 Hz), often linked to respiration and higher autonomic modulation \\
\hline
LF/HF Ratio & The ratio of LF to HF, indicating the balance between sympathetic and parasympathetic activity \\
\hline
LFn & Normalized LF, calculated by dividing LF by TP \\
\hline
HFn & Normalized HF, calculated by dividing HF by TP \\
\hline
LnHF & The natural logarithm of HF, often used to normalize the feature and improve its interpretability \\
\hline
\end{tabular}
\caption{Frequency-domain HRV features and their description.}
\label{tab:frequency_HRV_features}
\end{table}

\subsubsection{Electrodermal activity features}

Electrodermal activity (EDA) has been widely used as indicators of emotional and mental health states, including depression. EDA measures the variations in the electrical properties of the skin, resulting from the activity of sweat glands, recorded as changes in potential or variations in skin resistance. During emotional arousal, increased cognitive load, or physical activity, sweat production increases, leading to changes in the skin’s properties by raising conductivity and lowering resistance. Even a small amount of sweat, invisible to the naked eye on the skin's surface, can cause a noticeable change in its electrical conductivity \cite{ouzar2023reconnaissance}.

\begin{figure}[htbp]
\begin{center}
\includegraphics[width=8.5cm, height=5.5cm]{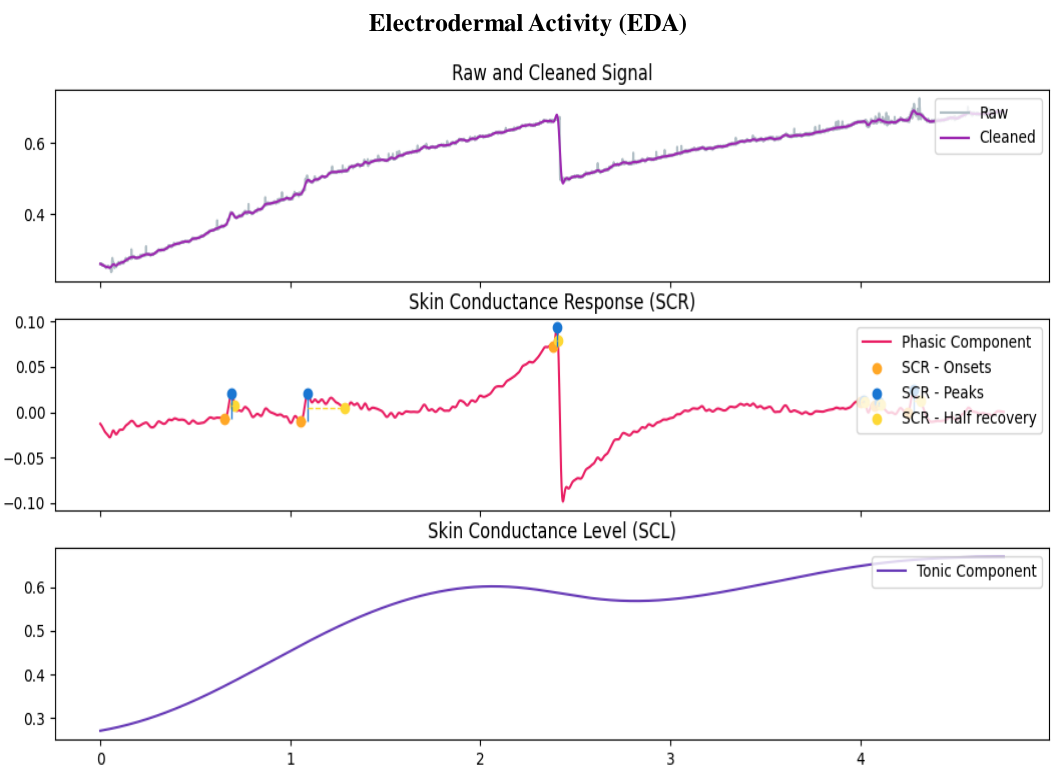}
\end{center}
   \caption{Visualization of an EDA signal. The top panel shows the raw and cleaned EDA signal. The middle panel displays the skin conductance response (SCR), highlighting the phasic component with identified SCR-onsets, SCR-peaks, and SCR-half recovery points. The bottom panel represents the skin conductance level (SCL), highlighting the tonic component of the EDA signal over time.}
\label{fig:eda}
\end{figure}

EDA signals were decomposed into two distinct components, namely phasic and tonic (see Figure \ref{fig:eda}). The tonic component represents the slow, continuous variations in electrodermal activity, capturing the baseline or background level of physiological arousal, such as chronic stress or anxiety. This component is extracted by isolating the low-frequency baseline variations in the raw signal. Once the tonic component is identified, it is subtracted from the raw EDA signal to obtain the phasic component. The phasic signal reflects the rapid and transient fluctuations in electrodermal activity, typically occurring in response to specific stimuli. These brief changes are linked to momentary activations of the sympathetic nervous system and are used to study stimulus-driven emotional or cognitive reactions. From each component, mean, standard deviation, minimum and maximum were computed as EDA features as well as the SCR amplitude of the signal excluding the tonic component ($SCR_\text{Amplitude}$) and the samples at which the onsets of the peaks occur ($SCR_\text{Onsets}$).

\subsubsection{Physical activity features}

Numerous cross-sectional studies have demonstrated a strong association between depression and increased sedentary behavior, likely attributed to reduced motivation and energy levels \cite{abd2023systematic}. However, the relationship between physical activity and depression is complex and can vary among individuals. 
To quantify physical activity, accelerometer data (ACC) is collected using the Empatica E4 wristband, which is equipped with a 3-axis accelerometer. This device captures movement in three dimensions (x, y, z) at a sampling frequency of 32 Hz, enabling the detection of activity patterns such as walking, running, resting, and even subtle motion characteristics.

Building upon the pre-processing steps performed in \cite{ghandeharioun2017objective}, the ACC data was processed using a 5th-order Butterworth low-pass filter with a cutoff frequency of 10 Hz. Subsequently, the filtered ACC data was transformed into motion features by computing the total accelerometer magnitude ($ACC_{Magnitude}$) which is the vector norm of the three-axis x, y and z using the following formula:

$ACC_{Magnitude} = \sqrt{ACC_x^2 + ACC_y^2 + ACC_z^2}$

where $ACC_x$, $ACC_y$, $ACC_z$ represent the accelerometer components in the x, y and z axes, respectively.

Key features extracted from accelerometer magnitude include measures of movement intensity (mean($ACC_{Mean}$), max($ACC_{Max}$), min($ACC_{Min}$), standard deviation ($ACC_{STD}$), and energy($ACC_{Energy}$)), movement dynamics (dominant frequency ($ACC_{Dominant-frequency}$), inactivity time($ACC_{Inactivity-time}$)), and movement symmetry (correlations between accelerometer signals on different axes ($Symmetry_ (x-y)$, $Symmetry_ (y-z)$, $Symmetry_ (x-z)$)).

The movement intensity measures ($ACC_{Mean}$, $ACC_{Max}$, $ACC_{Min}$, $ACC_{STD}$) are derived directly from the $ACC_{Magnitude}$ and correspond to its average, maximum, minimum, and standard deviation, respectively.

$ACC_{Energy}$ is defined as the sum of the squared magnitudes normalized by the ACC signal length : 

\[
ACC_\text{Energy} = \frac{\sum_{i=1}^{N} ACC_{Magnitude_i}^2}{N}
\]

where $N$ is the total number of samples. This feature quantifies the overall power of the accelerometer magnitude signal.

$ACC_{Dominant\_frequency}$ reflects the most significant periodic motion in the signal, often associated with repetitive activities such as walking or running. It is calculated using Fast Fourier Transform (FFT) and represents the frequency with the highest amplitude in the power spectrum. It is expressed as follows:
{\scriptsize
\[
ACC_\text{Dominant\_frequency} = 
 argmax ( | FFT(ACC_\text{Magnitude})| ) \times \frac{sampling\_rate}{length(ACC_\text{Magnitude})}
\]
}
where $FFT$ is the result of applying the FFT to the accelerometer magnitude signal. $argmax$ identifies the index of the maximum value in the magnitude of the FFT result. $sampling\_rate$ is the sampling frequency of the ACC data. $length(ACC_\text{Magnitude})$ is the length of the signal used in the FFT computation. $ACC_{Inactivity_time}$ quantifies periods with minimal or no movement, potentially indicating sedentary behavior or sleep. It is calculated using the following formula:
{\scriptsize
\[
ACC_\text{Inactivity\_time} =  \frac{\sum_{i=1}^{N} (ACC_{Magnitude_i} < inactivity-threshold)}{sampling\_rate}
\]
}
Symmetry between axes ($Symmetry_{x-y}$, $Symmetry_{y-z}$, $Symmetry_{x-z}$) quantifies the absolute correlations between ACC signals from different axes (X-Y, Y-Z, X-Z) and provides insights into the symmetry or alignment of movements across dimensions. Symmetry is computed as the absolute value of the Pearson correlation coefficient between the ACC signals of two axes, using the following formulas:
\[
\mathit{symmetry\_x\_y} = \left| \mathit{corr}(\mathit{signal\_x}, \mathit{signal\_y}) \right|
\]
\[
\mathit{symmetry\_y\_z} = \left| \mathit{corr}(\mathit{signal\_y}, \mathit{signal\_z}) \right|
\]
\[
\mathit{symmetry\_x\_z} = \left| \mathit{corr}(\mathit{signal\_x}, \mathit{signal\_z}) \right|
\]
where $corr$ is the pearson correlation coefficient.

\subsubsection{Temperature features}

Some previous research have revealed a significant correlation between depression and increased body temperature, suggesting that skin temperature could serve as a potential biomarker for this mental health condition \cite{lee2021current}. The Empatica E4 device measures peripheral skin temperature (TEMP) using an integrated optical thermometer. The TEMP data, recorded in degrees Celsius (°C), are sampled at a frequency of 4 Hz.

A set of statistical and temporal features are extracted from the TEMP signal to characterize its behavior and variability. These features include the mean temperature ($TEMP\_mean$), which represents the average value of the signal; the maximum ($TEMP\_max$) and minimum ($TEMP\_min$) temperatures, indicating the highest and lowest observed values; and the standard deviation ($TEMP\_std$), which measures the signal's variability around the mean. Additionally, the temperature range ($TEMP\_range$) captures the difference between the maximum and minimum values, while the trend ($TEMP\_trend$) quantifies the linear rate of change over time. The energy feature ($TEMP\_energy$) sums the squared deviations from the mean, reflecting overall signal variability.


\subsection{Proposed framework}

The overarching framework for depression subtypes—unipolar/bipolar- detection is illustrated in Figure \ref{framework}. We treat this task as a binary classification problem. The features extracted, as detailed in the previous section, serve as inputs for the classification step. Seven machine learning (ML) algorithms were applied and compared within our benchmark: Decision Tree (DT), Random Forest (RF), Gradient Boosting (GB), XGBoost (XGB), k-Nearest Neighbour (kNN), Support Vector Machine (SVM), and Multi Layer Perceptron (MLP). Since the entire data processing pipeline was implemented in Python, we utilized the scikit-learn library's implementations of the aforementioned classifiers. Due to the absence of publicly available datasets specifically designed for detecting depression subtypes and providing wearable signals, we employed the Leave-One-Out Cross-Validation (LOOCV) strategy to evaluate model performance using our own dataset, as described in section \ref{dataset}. To optimize the performance of the ML algorithms, we implemented a grid search strategy to systematically explore and evaluate various hyperparameter combinations, thereby identifying the optimal configuration for each algorithm.

\section{RESULTS AND DISCUSSION}

Two different experiments were performed for depression subtypes classification using wearable signals collected using Empatica E4, both separately and in combination. The analysis leveraged physiological and behavioral data, namely, heart rate variability features in time and frequency domain, skin temperature features, electrodermal and physical activity. To comprehensively evaluate the classification performance of the ML models using the LOOCV strategy, the following metrics were employed: accuracy, precision, recall, and F1 score. These metrics are defined as follows:

\begin{enumerate}

\item {Accuracy: This metric measures the proportion of correct predictions (true positives and true negatives) out of the total instances. It is a useful metric for evaluating overall model performance, but it can be misleading, especially in imbalanced datasets. If one class significantly outnumbers the other, the model can achieve high accuracy simply by predicting the majority class for all instances. It is calculated as:

\begin{equation}\scriptsize
    Accuracy = \frac{True Positives (TP) + True Negatives (TN)}{TP + TN + False Positives + False Negatives}
\label{equation:accuracy}
\end{equation}
}

\item {Precision: This metric measures the proportion of correct predictions that are actually correct (true positives) among all positive predictions made by the model. It reflects how accurate the model is in avoiding false positives (classifying for example unipolar depression as bipolar). It is defined as:  

\begin{equation}\scriptsize
    Precision = \frac{True Positives}{True Positives + False Positives}
\label{equation:precision}
\end{equation}
}

\item {Recall: Recall, also known as sensitivity or true positives rate, measures the proportion of true positives that are correctly identified by the model out of all actual positive instances. It reflects how effective the model is in avoiding false negatives. It is calculated as:

\begin{equation}\scriptsize
    Recall = \frac{True Positives}{True Positives + False Negatives}
\label{equation:Recall}
\end{equation}
}

\item {F1-score: The F1-score is defined as the harmonic mean of the model’s precision and recall. It provides a balanced measure of the model's performance, especially useful when dealing with imbalanced datasets. It takes into account both the ability to correctly identify true positives (recall) and the ability to avoid false positives (precision). It is calculated as:

\begin{equation}\scriptsize
    F1-score = \frac{2*Precision*Recall}{Precision+Recall} 
\label{equation:F1-score}
\end{equation}
}

\end{enumerate}

\setlength{\arrayrulewidth}{0.3mm}
\setlength{\tabcolsep}{4pt}
\renewcommand{\arraystretch}{1.2}

\begin{table}[htbp]
\caption{\label{table:phys}Unipolar Vs Bipolar depression classification results based on time-domain HRV features.}
\centering
\small

\begin{tabular}{|c|c|c|c|c|}

    \hline
    \textbf{Method} & \textbf{Accuracy} & \textbf{Precision} & \textbf{Recall} & \textbf{F1 Score} \\
    \hline
    Gradient Boosting & 48,39 & 54,55 & 66,67 & 60 \\
    \hline
    RF & 48,39 & 54,16 & 72,22 & 61,90 \\
    \hline
    kNN & 51,61 & 57,14 & 66,67 & 61,54 \\
    \hline
    DT & 58,06 & 64,71 & 61,11 & 62,86 \\
    \hline
    XGBOOST & 61,29 & 60,71 & 94,44 & 73,91 \\
    \hline
    SVM & 61,29 & 60 & \textbf{100} & 75 \\
    \hline
    MLP & \textbf{77,42} & \textbf{72} & \textbf{100} & \textbf{83,72} \\
    \hline
\end{tabular}
\label{time_HRV}

\end{table}

\subsection{Performance analysis of individual wearable signals}

The average of each metric across all subjects is reported in Tables \ref{time_HRV}, \ref{frequency_HRV}, \ref{eda}, \ref{acc}, and \ref{temp}. These tables present the performance metrcis of various machine learning methods for classifying subtypes of depression using five distinct physiological and behavioral signals evaluated separately: time-domain HRV features, frequency-domain HRV features, electrodermal activity features, accelerometer features, and skin temperature features, respectively.

For time-domain HRV features, the MLP significantly outperforms all others ML models, achieving the highest accuracy (77.42\%), precision (72\%), recall (100\%), and F1 score (83.72\%). In contrast, GB, RF, and kNN performed poorly, achieving an accuracy $\leq$ 48,39\%.

\setlength{\arrayrulewidth}{0.3mm}
\setlength{\tabcolsep}{4pt}
\renewcommand{\arraystretch}{1.2}

\begin{table}[htbp]
\caption{\label{table:phys}Unipolar Vs Bipolar depression classification results based on frequency-domain HRV features.}
\centering
\small

\begin{tabular}{|c|c|c|c|c|}

    \hline
    \textbf{Method} & \textbf{Accuracy} & \textbf{Precision} & \textbf{Recall} & \textbf{F1 Score} \\
    \hline
    Gradient Boosting & 48,39 & 55 & 61,11 & 57,89 \\
    \hline
    RF & 48,39 & 54,55 & 66,67 & 60 \\
    \hline
    DT & 54,84 & 59,09 & 72,22 & 65 \\
    \hline
    XGBOOST & 61,29 & \textbf{68,75} & 61,11 & 64,71 \\
    \hline
    kNN & 61,29 & 65 & 72,22 & 68,42 \\
    \hline
    SVM & 64,52 & 62,96 & \textbf{94,44} & 75,56 \\
    \hline
    MLP & \textbf{67.74} & 66.67 & 88.89 & \textbf{76.19} \\
    \hline
\end{tabular}
\label{frequency_HRV}

\end{table}

Frequency-domain HRV features, across all ML models, show modest performance compared to others wearable signals, with accuracies ranging from approximately 48\% to 67\%. MLP again led with 67.74\% accuracy and 76.19\% F1 Score, though its performance slightly declined compared to time-domain HRV results. SVM mirrors the earlier precision-recall trade-off, achieving 94.44\% recall but only 62.96\% precision. DT and RF struggle (F1 $\leq$ 65\%), while kNN and XGB show moderate performance (61.29\% accuracy).

\setlength{\arrayrulewidth}{0.3mm}
\setlength{\tabcolsep}{4pt}
\renewcommand{\arraystretch}{1.2}

\begin{table}[htbp]
\caption{\label{table:phys}Unipolar Vs Bipolar depression classification results based on EDA features.}
\centering
\small

\begin{tabular}{|c|c|c|c|c|}

    \hline
    \textbf{Method} & \textbf{Accuracy} & \textbf{Precision} & \textbf{Recall} & \textbf{F1 Score} \\
    \hline
    kNN & 51,61 & 57,89 & 61,11 & 59,46 \\
    \hline
    RF & 58,06 & 61,90 & 72,22 & 66,67 \\
    \hline
    Gradient Boosting & 74,19 & 77,78 & 77,78 & 77,78 \\
    \hline
    SVM & 74,19 & 77,78 & 77,78 & 77,78 \\
    \hline
    DT & 77,42 & \textbf{78,95} & 83,33 & 81,08 \\
    \hline
    XGBOOST & 77,42 & 76,19 & 88,89 & 82,05 \\
    \hline
    MLP & \textbf{83.87} & 78.26 & \textbf{100} & \textbf{87.80} \\
    \hline
\end{tabular}
\label{eda}

\end{table}

EDA features proved more discriminative than time and frequency domain HRV features, with MLP dominating (83.87\% accuracy, 87.80\% F1 Score) alongside perfect recall. XGB and DT also perform well, both achieving 77.42\% accuracy, while GB and SVM exhibit similar results (74.19\% accuracy, 77.78\% F1 Score).

The classification results using ACC data, as presented in Table \ref{acc}, indicate that features related to physical activity are the most effective in distinguishing between unipolar and bipolar depression, outperforming other physiological signals in terms of accuracy and overall model performance. MLP achieves the highest overall performance with an accuracy of 96,77\%, precision of 100\%, recall of 94,44\%, and an F1 score of 97,14\%. Other methods like GB and XGB also exhibit good results, with high recall scores (up to 100\%), indicating their ability to correctly predicting all instances of one of the two classes. In contrast, NB and kNN show modest performance, with accuracy ranging from 50-60\%. 

\setlength{\arrayrulewidth}{0.3mm}
\setlength{\tabcolsep}{4pt}
\renewcommand{\arraystretch}{1.2}

\begin{table}[htbp]
\caption{\label{table:phys}Unipolar Vs Bipolar depression classification results based on ACC features.}
\centering

\small

\begin{tabular}{|c|c|c|c|c|}

    \hline
    \textbf{Method} & \textbf{Accuracy} & \textbf{Precision} & \textbf{Recall} & \textbf{F1 Score} \\
    \hline
    kNN & 48,39 & 66,67 & 22,22 & 33,33 \\
    \hline
    RF & 64,52 & 68,42 & 72,22 & 70,27 \\
    \hline
    DT & 64,52 & 68,42 & 72,22 & 70,27 \\
    \hline
    SVM & 67,74 & 64,29 & \textbf{100} & 78,26 \\
    \hline
    XGBOOST & 77,42 & 72 & \textbf{100} & 83,72 \\
    \hline
    Gradient Boosting & 80,65 & 77,27 & 94,44 & 85 \\
    \hline
    MLP & \textbf{96.77} & \textbf{100} & 94.44 & \textbf{97.14} \\
    \hline
\end{tabular}
\label{acc}

\end{table}

Finally, TEMP features further highlighted MLP’s robustness, achieving excellent performance with 93.55\% accuracy, 90\% precision, 100\% recall, and a 94.74\% F1 Score. XGB (77.42\% accuracy) and kNN (70.97\% accuracy) also showed good results. SVM, on the other hand, prioritized recall (100\%) over precision (60\%), successfully identifying all instances of one of the two classes but generating a significant number of false positives. 

\setlength{\arrayrulewidth}{0.3mm}
\setlength{\tabcolsep}{4pt}
\renewcommand{\arraystretch}{1.2}

\begin{table}[htbp]
\caption{\label{table:phys}Unipolar Vs Bipolar depression classification results based on TEMP features.}
\centering
\small

\begin{tabular}{|c|c|c|c|c|}

    \hline
    \textbf{Method} & \textbf{Accuracy} & \textbf{Precision} & \textbf{Recall} & \textbf{F1 Score} \\
    \hline
    RF & 61,29 & 66,67 & 66,67 & 66,67 \\
    \hline
    DT & 61,29 & 66,67 & 66,67 & 66,67 \\
    \hline
    SVM & 61.29 & 60 & 100 & 75 \\
    \hline
    Gradient Boosting & 67,74 & 70 & 77,78 & 73,68 \\
    \hline
    kNN & 70,97 & 80 & 66,67 & 72,73 \\
    \hline
    XGBOOST & 77,42 & 78,95 & 83,33 & 81,08 \\
    \hline
    MLP & \textbf{93.55} & \textbf{90} & \textbf{100} & \textbf{94.74} \\
    \hline
\end{tabular}
\label{temp}

\end{table}

\setlength{\arrayrulewidth}{0.3mm}
\setlength{\tabcolsep}{4pt}
\renewcommand{\arraystretch}{1.2}

\begin{table}[htbp]
\caption{\label{table:phys}Unipolar Vs Bipolar depression classification results based on combined physiological and behavioal features.}
\centering
\small

\begin{tabular}{|c|c|c|c|c|}

    \hline
    \textbf{Method} & \textbf{Accuracy} & \textbf{Precision} & \textbf{Recall} & \textbf{F1 Score} \\
    \hline
    RF & 45.16 & 52.17 & 66.67 & 58.54 \\
    \hline
    SVM & 58.06 & 58.06 & \textbf{100} & 73.47 \\
    \hline
    kNN & 64.52 & 73.33 & 61.11 & 66.67 \\
    \hline
    Gradient Boosting & 70.97 & 73.68 & 77,78 & 75.68 \\
    \hline
    XGBOOST & 74,19 & 72.73 & 88.89 & 80 \\
    \hline
    DT & 77.42 & 86.67 & 72.22 & 78.79 \\
    \hline
    MLP & \textbf{93.55} & \textbf{100} & 88.89 & \textbf{94.12} \\
    \hline
\end{tabular}
\label{phys_beh}

\end{table}

\subsection{Performance analysis of combined wearable signals}

The last table \ref{table:phys} shows the performance achieved by combining all wearable-derived physiological and behavioral features.
The results indicate that while the MLP remained the top-performing model, achieving 93.55\% accuracy and a 94.12\% F1 score, this performance slightly underperformed compared to using individual ACC (96.77\% accuracy) or TEMP (93.55\% accuracy) features. This suggests potential issues with redundancy, noise, or irrelevant interactions in the combined data.
Other models, such as XGB and GB, achieve moderate accuracy (74.19–74.94\%) and F1 scores (77.78\%), while SVM maintained its trend of prioritizing recall (100\%) over precision (58.06\%), leading to a lower F1 score (73.47\%). Notably, simpler models like kNN show improved performance (70.97\% accuracy, 72.73\% F1) compared to their results with individual HRV or ACC features, implying that combined features may partially compensate for weaker individual modalities in some cases.

\subsection{Discussion}

The analysis of individual wearable signals revealed notable differences in performance between physiological and behavioral markers. Features derived from accelerometer data, representing physical activity, achieved the highest classification accuracy, reaching 97\% with MLP model, suggesting that behavioral markers may be more effective for discriminating between unipolar and bipolar depression. This could reflect inherent differences in motor activity, energy levels, or circadian rhythm disruptions between the two disorders. Temperature-related features followed, with an accuracy of 93\%, while electrodermal activity features showed promising but slightly lower performance compared to behavioral modalities. In contrast, HRV features extracted from BVP signals demonstrated the weakest performance, particularly in the frequency domain. The relatively low performance of physiological features could be attributed to challenges in signal quality and noise, as BVP signals are prone to motion artifacts, environmental interference, and variability in sensor contact during real-world data collection. These limitations could obscure subtle physiological differences between unipolar and bipolar depression, reducing the reliability of derived biomarkers. While EDA features exhibited moderate success, their performance likely depends on high-quality signal acquisition and pre-processing to mitigate noise from factors like sweat or ambient temperature.
Notably, the lack of significant improvement with combined features underscores the challenges of multimodal data integration. While combining features from multiple modalities theoretically offers the potential to capture complementary information, the results suggest that overlapping or redundant information between physiological and behavioral markers may reduce the effectiveness of such combinations. This highlights the importance of careful feature engineering and the development of advanced fusion techniques to effectively leverage the synergies between modalities.

\section{CONCLUSIONS AND FUTURE WORKS}

In this paper, we presented a novel framework for continuous, passive, and non-invasive detection of depression subtypes (unipolar vs. bipolar) based on objective markers extracted from wearable device signals. We proposed also a new multimodal depression dataset named CALYPSO, which contains physiological and behavioral data collected via Empatica E4, alongside video and audio recordings from patients diagnosed with severe depression. A comprehensive set of well-known time-domain, frequency-domain and statistical features were extracted from the wearable signals and evaluated separately and combined using common machine learning algorithms. Experimental results showed that physical activity features extracted from accelerometer was the most important for discriminating between unipolar and bipolar disorder followed by temperature features.

Building on our work with wearable sensor data, future research will explore a multimodal approach, integrating wearable-derived physiological and behavioral data with video and audio based behavioral cues to capture a more comprehensive set of affective signals. This will involve analyzing vocal tone, facial expressions, head movements, and gestures. Furthermore, we aim to extend our work by exploring the detection of other depression subtypes beyond unipolar and bipolar depression, such as melancholic, atypical, or anxious depression, as well as identifying their characteristics. An in-depth analysis of individual features from each wearable signal will also be conducted to better understand the qualitative aspects of the data. This analysis may help uncover why combining features does not consistently yield significant improvements over the best-performing individual features, providing insights for more targeted and effective multimodal feature integration strategies.


{\small

}


\end{document}